\documentclass[lettersize,journal]{IEEEtran}
\usepackage{amsmath,amsfonts}
\usepackage{algorithmic}
\usepackage{algorithm}
\usepackage{float}
\usepackage{array}
\usepackage{times}
\usepackage[caption=false,font=footnotesize,labelfont=rm,textfont=rm]{subfig}
\usepackage{textcomp}
\usepackage{stfloats}
\usepackage{makecell}
\usepackage{url}
\usepackage{verbatim}
\usepackage{graphicx}
\usepackage{epsfig}
\usepackage{epstopdf} 
\usepackage{cite}
\usepackage{multirow}
\begin{document}
\title{Diagnosis of Fuel Cell Health Status with Deep Sparse Auto-Encoder Neural Network}

\author{Chenyan Fei, Dalin Zhang, Chen Melinda Dang\\   Hangzhou Dianzi University, Shanghai High School International Division\\241080052@hdu.edu.cn, zhangdalin90@gmail.com, Melinda\_cd@outlook.com }




\maketitle

\begin{abstract}
Effective and accurate diagnosis of fuel cell health status is crucial for ensuring the stable operation of fuel cell stacks. Among various parameters, high-frequency impedance serves as a critical indicator for assessing fuel cell state and health conditions. However, its online testing is prohibitively complex and costly. This paper employs a deep sparse auto-encoding network for the prediction and classification of high-frequency impedance in fuel cells, achieving metric of accuracy rate above 92\%. The network is further deployed on an FPGA, attaining a hardware-based recognition rate almost 90\%.
\end{abstract}


\section{Introduction}
\IEEEPARstart{U}{nder} the mission of zero carbon emissions, Electric Vehicles (EV) have been on the rise in the automotive industry, forming part of a global movement toward sustainability. It is well-known that the battery makes up a significant portion of the cost and environmental impact of an EV, so the lifetime of the Fuel Cells (FC) become a critical factor.

Diagnosing the health status and maintaining the overall efficiency of a fuel cell stack is vitally important \cite{ref1}. Internal faults may arise, leading to catalyst degradation. This reduces activity and thus degrading energy conversion efficiency and power output. Direct measurement of FC health state is prohibitively complex and costly using electrochemical impedance spectroscopy (EIS), several approaches have been investigated to forecast the degradation of FCs: model-based methods, data-based methods, and hybrid methods \cite{ref1,ref2,ref3}. However, the model-based electrochemical mechanisms within FCs are complex, involving multi-time scale aging phenomena and multi-physical domain interactions \cite{ref4}. Hence, such complexity makes it yet a challenge to build a reliable model \cite{ref5}. Moreover, model-based approach often is not adaptive to differing operation conditions, failing to generalize the internal mechanisms of FCs, limiting the accuracy of such models. Therefore, data-driven methods have been widely adopted. These include (1) non-parametric regression \cite{ref6}, (2) probabilistic/statistical methods \cite{ref7}, and (3) neural networks, including long short-term memory (LSTM) network \cite{ref8}, which are well suited for time series prediction because of their long-term memory capabilities; Or Gaussian process regression methods \cite{ref9}, or general machine learning algorithms \cite{ref10}. 

While FC health diagnosis and remaining useful life (RUL) prediction using deep learning / probabilistic models are common, they result in low accuracy. Traditional neural network and classical machine learning models often struggle to deal with noisy raw data and may fail to capture global nonlinear trends. In addition, some machine learning techniques, such as XG Boost, LSTM are prone to over-fitting in small or noisy datasets. To address these, Deep Sparse Auto-Encoder (DSAE) is proposed as a multi-layer neural architecture for sequential data: by penalizing neuron activation, it can reduce overfitting and better capture latent features.  

On the implementation side, a recent study from Tongji University \cite{ref11} has successfully implemented the system using a micro-chip controller. A multilayer perceptron (MLP) neural network is employed for the prediction of FC high-frequency impedance and deployed on Infineon AURIX TC499 platform. However, this method could not be generalized to different scenes. In this work, a Field Programmable Gate Arrays (FPGA)-based implementation is proposed using Xilinx hardware to ensure computational efficiency and applicability in real‐time or embedded settings.

The outline of this work is as follows. In Section I, the background of the FC health diagnosis task is introduced. Next, the DSAE model is proposed in detail in Section II. Section III shows simulation results of the DSAE model in Python. Section IV discusses the FPGA implementation. Finally, Section V concludes the paper.
\IEEEpubidadjcol
\section{Overview of the DSAE Network Model}
\begin{figure}[H]
	\vspace{-0.35cm}
	\centering
	\includegraphics[width=3.1in]{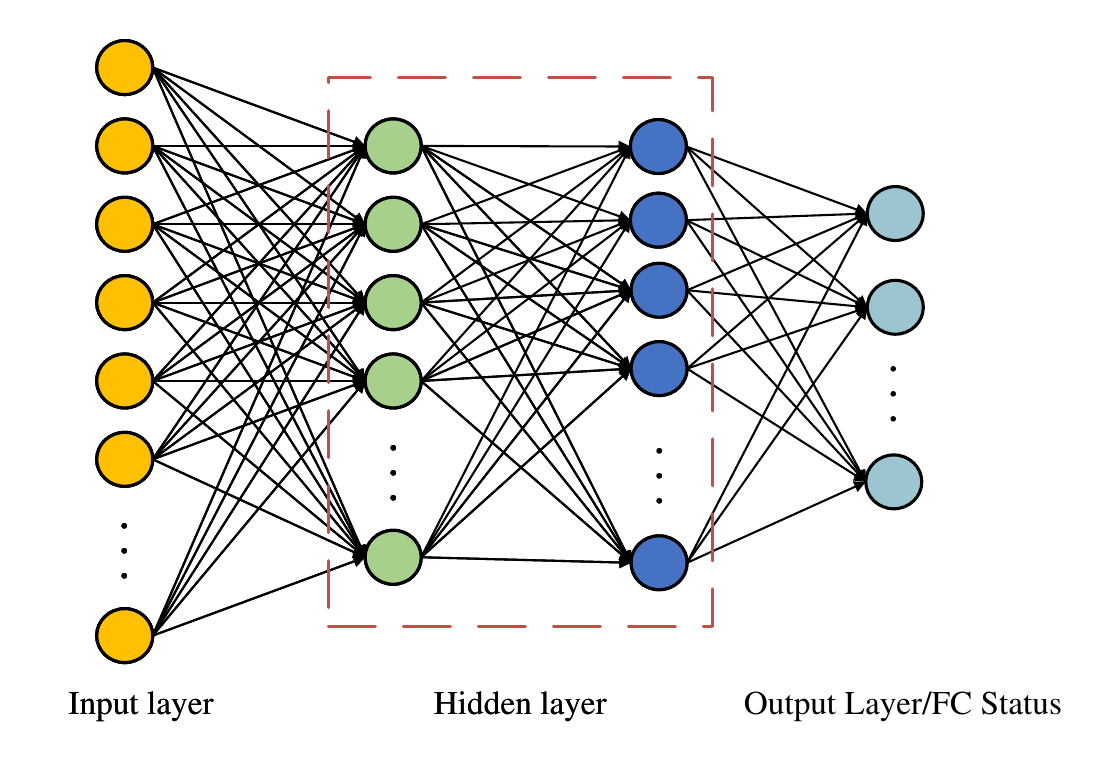}
	\caption{The structure of DSAE network}
	\label{Fig1}
\end{figure}

The DSAE network for FC diagnosis, based on the sparse auto-encoder, exhibits a gradually decreasing loss function between its output values and input samples. It achieves feature compression with lower computational complexity and a reduced parameter count. The structure is illustrated in Fig. \ref{Fig1}.
\IEEEpubidadjcol

The DSAE network imposes sparsity constraints on the neurons in the hidden layer by adding a penalty term to the loss function. The Kullback-Leibler divergence is adopted for sparsity constraint, and the mathematical model is as follows:
\begin{equation}
	KL\left( \xi \left\| {{\xi }_{k}} \right. \right)=\xi \log \frac{\xi }{{{\xi }_{k}}}+\left( 1-\xi  \right)\log \frac{1-\xi }{1-{{\xi }_{k}}},
	\label{11}
\end{equation}
\begin{equation}
	{{\xi }_{k}}=\frac{1}{p}\sum\limits_{i=1}^{p}{\left[ {{h}_{k}}({{x}_{i}}) \right]},
	\label{22}
\end{equation}
where $\xi $ is the sparsity coefficient; ${{\xi }_{k}}$ is the average activation degree of the \textit{k}-th neuron in the hidden layer; ${{h}_{k}}({{x}_{i}})$ represents the activation degree of the \textit{k}-th neuron in the hidden layer under input ${{x}_{i}}$. The total loss function after adding sparse constraints is:
\begin{equation}
	{{J}_{total}}=J(\textit{\textbf{W}},\textit{\textbf{b}})+\psi \sum\limits_{k=1}^{q}{KL\left( \xi \left\| {{\xi }_{ik}} \right. \right)},
	\label{33}
\end{equation}
where $\psi$ represents sparse weight factor; \textit{q} is the number of hidden layer neurons. $\textit{\textbf{W}}$ and $\textit{\textbf{b}}$ represent each layer weight and bias. $J(\textit{\textbf{W}},\textit{\textbf{b}})$ is the mean squared error loss function.
\begin{figure}[H]
	\vspace{-0.35cm}
	\centering
	\includegraphics[width=2.9in]{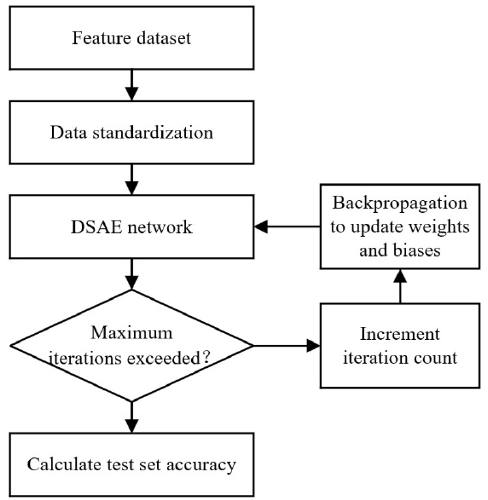}
	\caption{The training process of DSAE network}
	\label{Fig2}
\end{figure}
The overall training process of DSAE network is illustrated in Fig. \ref{Fig2}. The training procedure begins with standardizing the constructed sample datasets, which is then fed into the DSAE network. The output values of the network are compared with the true labels to compute the loss function value for training iteration. When the current iteration count is below the maximum, backpropagation is performed to update the weights and biases of each layer, thereby reducing the loss function value and improving the network's recognition performance. This process continues until the preset maximum iteration count is reached, after which the accuracy of the test set is calculated.
\section{ Prediction and Classification Based on DSAE Network for High-Frequency Impedance Values}
\subsection{Description of the Dataset}
- Data source: Collected from fuel cell OEM test benches

- Data composition: Includes time, neural network inputs (output power, current density, stack voltage, variance of single-cell voltage, outlet water temperature, hydrogen pressure, proportional valve current, hydrogen circulation pump power, air pressure, air flow), and the output predictions is high-frequency resistance (HFR) or impedance.

- Operating conditions: Continuous operation for over 10 hours based on cyclic conditions. Each cycle includes low, medium, and high loads, as well as load increase and decrease scenarios.

\begin{table*}[bhtp]
	\caption{Sample Records from the Dataset}
	\label{T1}
	\centering
	\tabcolsep=0.025\linewidth
	\begin{tabular}{c c c c c c}
		\hline
		t($\text{s}$) & Power($\text{kW}$)& CurrD($\text{mA/cm}^{2}$)&StaVol($\text{V}$) &Var&WaterTempOut($^{\circ}\text{C}$)\\
		\hline
		1 & 24.2 &222.4 &363.8 &83 &68.5 \\
		2 & 24 &222.5 &364.1 &74 &68.6\\
		3 & 24.2 &222.3 &364 &73 &68.6 \\
		4 &23.9 &222.7 &363.8 &75 &68.6 \\
		\hline
	\end{tabular}
\hspace{2em}
		\begin{tabular}{c c c c c}
		\hline
		H2PressIn($\text{kPaG}$) & HCPPower& AirPressIn($\text{(kPaG}^{2}$)&AirFlow($\text{g/s}$) &HFR($\text{m}\Omega$)\\
		\hline
        165.5 & 0.44 & 145.6 &28.6 &88.5\\
        165.6 & 0.43 & 145.7 &29 &88\\
        165.4 &0.42 &145.6 &27.8 & 88\\
        165.8 &0.41 &145.6 &29 & 88\\
		
		\hline
	\end{tabular}
\end{table*}

Table \ref{T1} displays several sample records from the dataset. The classification criteria are defined as follows: Class 0 corresponds to HFR $<$ 89, Class 1 corresponds to 89 $\leq$ HFR $<$ 91, and Class 2 corresponds to HFR $\geq$ 91. The complete dataset comprises 36363 records, which are partitioned into training and testing sets at a 3:1 ratio for input into the classification network to facilitate model training and validation.
\subsection{Structure and Parameter Configuration of the DSAE Network}
The DSAE network is configured with an input layer containing 10 nodes and an output layer with 3 nodes, corresponding to the number of modulation categories. The network comprises two hidden layers with 32 and 16 neurons respectively, utilizing the ReLU activation function for both hidden and output layers. The Adam optimizer is employed with a learning rate of 0.001.
\subsection{DSAE Network Classification Results Based on Python}
\begin{figure}[H]
	\vspace{-0.35cm}
	\centering
	\includegraphics[width=2.9in]{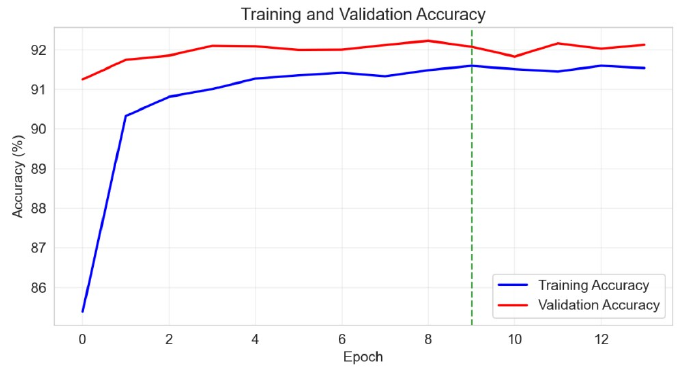}
	\caption{Training and validation accuracy}
	\label{Fig3}
\end{figure}

\begin{figure}[H]
	\vspace{-0.35cm}
	\centering
	\includegraphics[width=2.9in]{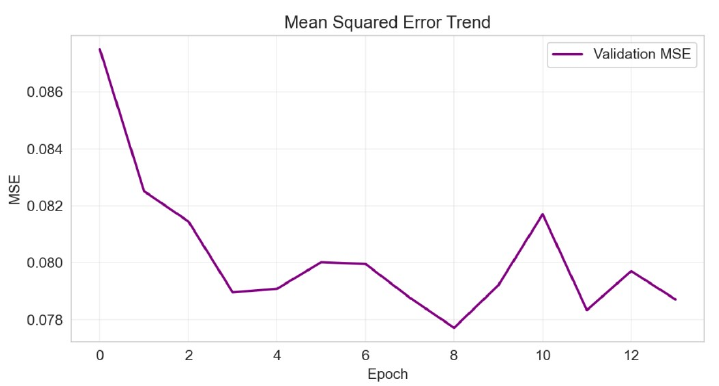}
	\caption{Mean squared error trend}
	\label{Fig4}
\end{figure}

\begin{figure}[H]
	\vspace{-0.35cm}
	\centering
	\includegraphics[width=2.9in]{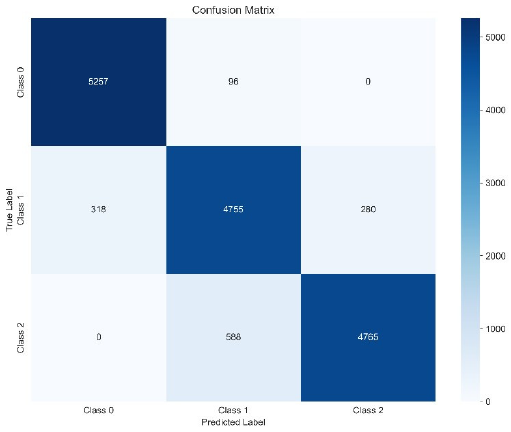}
	\caption{The confusion matrix}
	\label{Fig5}
\end{figure}

The simulations are conducted in Python within the PyCharm 2024.3.3 environment. The training results of the DSAE network are presented in Table \ref{T2}, achieving a classification accuracy of 92.13\% with a mean squared error (MSE) of 0.0787. Other evaluation metrics, including precision, recall, and F1-score, all exceeded 90\%.
\begin{table}[H]
	\caption{Training results of DSAE network}
	\label{T2}
	\centering
	\tabcolsep=0.04\linewidth
	\begin{tabular}{c c c }
		\hline
		Indicators& & Value \\
		\hline
        Accuracy&&	0.9213\\
        Precision&&	0.9221\\
        Recall&&	0.9213\\
        F1-Score&&0.9214\\
        MSE&&0.0787\\
        
		\hline
	\end{tabular}
\end{table}
During the training process, the training and validation accuracy are shown in Fig. \ref{Fig3}. The optimal epoch is 9, with the training accuracy exceeding 90\% once the epoch count surpassed 3. Fig. \ref{Fig4} shows the trend of MSE, where it can be observed that the MSE consistently remains below 0.1.
The confusion matrix of the classification results is shown in Fig. \ref{Fig5}. The values are primarily concentrated along the diagonal, with the model demonstrating the best recognition performance for Class 1.

\section{FPGA Implementation}
\subsection{IP Core Generation}
\begin{figure}[H]
	\vspace{-0.35cm}
	\centering
	\includegraphics[width=3.3in]{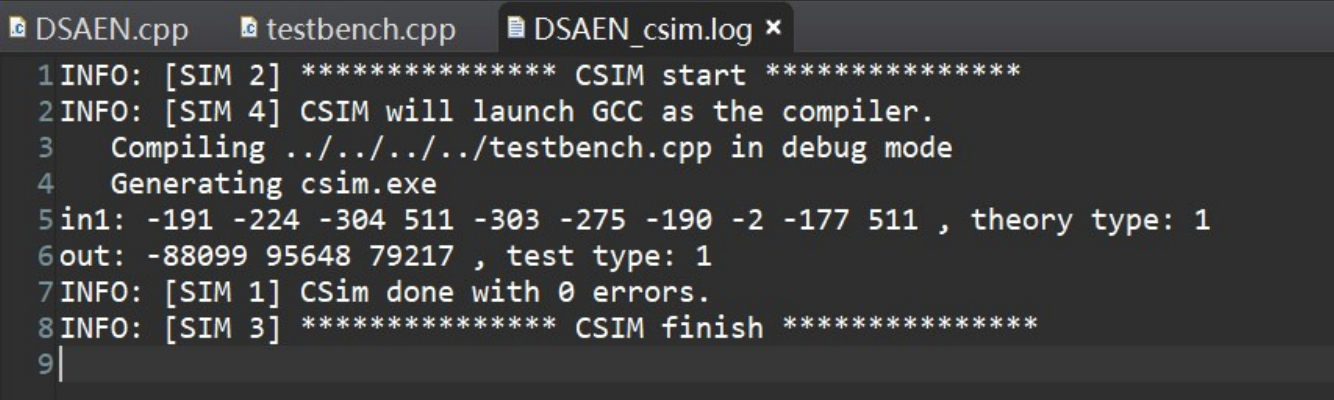}
	\caption{Functional validation results}
	\label{Fig61}
\end{figure}

\begin{figure}[H]
	\vspace{-0.35cm}
	\centering
	\includegraphics[width=2.8in]{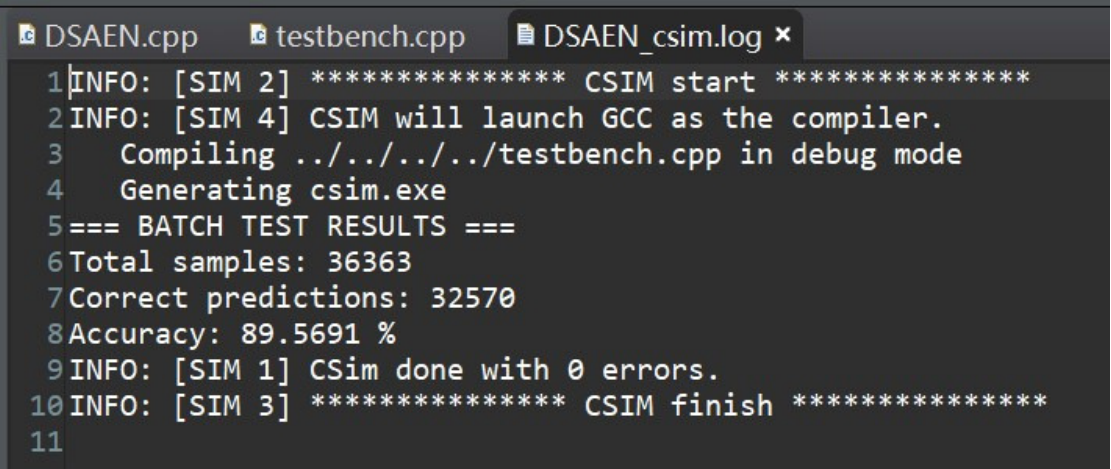}
	\caption{Accuracy evaluation results}
	\label{Fig62}
\end{figure}

\begin{figure}[H]
	\vspace{-0.35cm}
	\centering
	\includegraphics[width=2.2in]{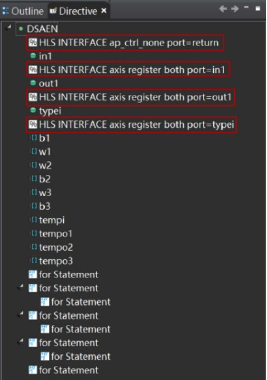}
	\caption{Directive configuration}
	\label{Fig7}
\end{figure}

\begin{figure}[H]
	\vspace{-0.35cm}
	\centering
	\includegraphics[width=2.8in]{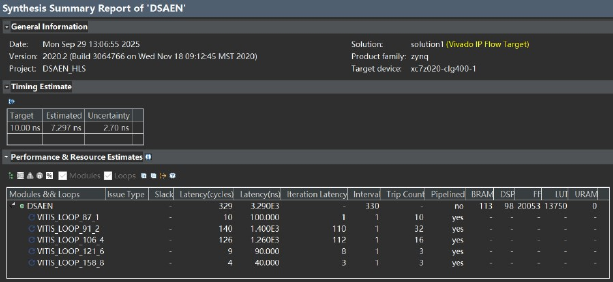}
	\caption{Synthesis performance summary}
	\label{Fig8}
\end{figure}

\begin{figure}[H]
	\vspace{-0.35cm}
	\centering
	\includegraphics[width=2.8in]{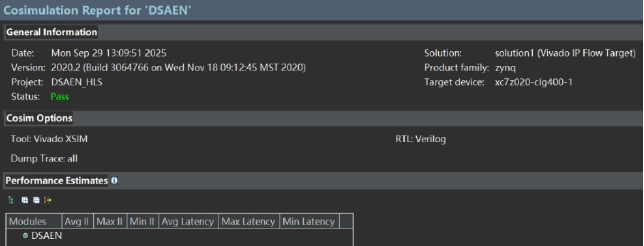}
	\caption{C/RTL simulation report}
	\label{Fig9}
\end{figure}

\begin{figure}[H]
	\vspace{-0.35cm}
	\centering
	\includegraphics[width=2.2in]{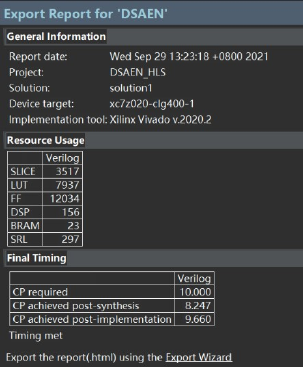}
	\caption{IP generation report}
	\label{Fig10}
\end{figure}

The hardware IP core is developed using Vivado HLS 2020.2 for the PYNQ-Z2 board. The core algorithm is implemented in C++ and validated through simulation. A single test case of class 0 confirmed functional correctness, which is shown in Fig. \ref{Fig61}, while comprehensive testing across the dataset achieved 89.569\% accuracy, which is shown in Fig. \ref{Fig62}. Thus, performance baseline is established.
Appropriate directives are then applied to the top-level function and interface parameters for hardware optimization, as illustrated in Fig. \ref{Fig7}. The subsequent synthesis process generates detailed reports, shown in Fig. \ref{Fig8}, providing general information, timing estimate, and performance \& resource estimates. Moreover, the generated RTL code is verified by C/RTL co-simulation, as demonstrated in Fig. \ref{Fig9}. Finally, the design is exported as a packaged IP core, shown in Fig. \ref{Fig10}.
\subsection{System Integration}
\begin{figure*}[t!]
	\centering
	\includegraphics[width=0.9\textwidth]{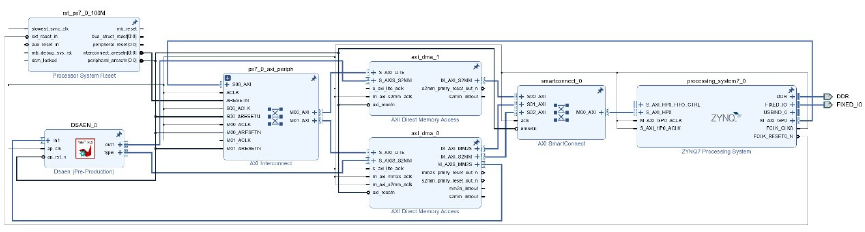} 
	\caption{Block design}
	\label{Fig11}
\end{figure*}

The generated IP is integrated into a Vivado project. An AXI DMA controller is used to facilitate high-speed data transfer between the PS and the PL, resulting in the complete block design shown in Fig. \ref{Fig11}. After synthesis and implementation, the bitstream is generated. The necessary deployment files, including .tcl, .bit, and .hwh formats, are then exported to a designated folder on the PYNQ board.
\subsection{Functional Verification}
\begin{figure}[H]
	\vspace{-0.35cm}
	\centering
	\includegraphics[width=2.2in]{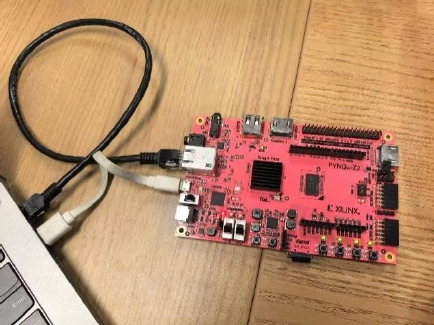}
	\caption{PYNQ-Z2 connection diagram}
	\label{Fig12}
\end{figure}

\begin{figure*}[t!]
	\centering
	\includegraphics[width=0.9\textwidth]{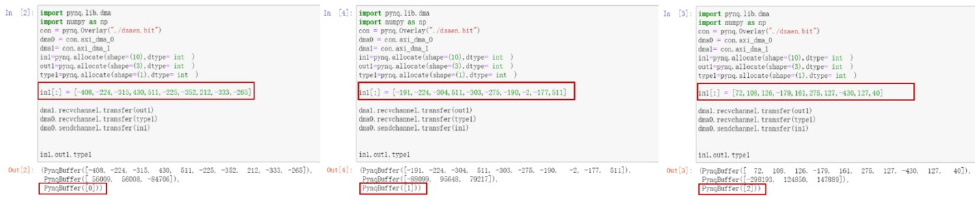} %
	\caption{Classification results for class labels 0, 1, and 2}
	\label{Fig13}
\end{figure*}

The hardware system is deployed and tested on the PYNQ-Z2 board as shown in Fig. \ref{Fig12}. Hardware acceleration is driven by Python scripts within the Jupyter Notebook environment. As shown in Fig. \ref{Fig13}, classification results for sample inputs among classes 0, 1, and 2 matched both the dataset labels and the prior simulation results via Python. This validates the hardware implementation, which achieved a classification accuracy of 89.569\%. 
\section{Conlusion}
The DSAE network demonstrates significant effectiveness in classifying the high-frequency impedance of fuel cells. The model achieves the metric of accuracy rate above 92\%. Furthermore, the deployment of the network on the FPGA platform highlights its practical applicability, achieving a hardware recognition rate of almost 90\%. This implementation not only provides a feasible solution for online monitoring of fuel cells but also offers a cost-effective alternative to complex and expensive traditional testing methods.
\section*{Acknowledgements}
The authors are indebted to Professor Yuan Zhu from Tongji University for his guidance on this research work.

 

\begin{thebibliography}{1}
\bibliographystyle{IEEEtran}





\bibitem{ref1}
Zhao Lei, et al. "A novel pre-diagnosis method for health status of proton exchange membrane fuel cell stack based on entropy algorithms."eTransportation 18. 2023: doi:10.1016/J. ETRAN.2023.100278.

\bibitem{ref2}
L. Fengyang, Y. Tian, Y. Zhanhui, W. Jia and W. Zilong, "Research on Data Efficiency of Fuel Cell Vehicles Based on Demonstration Platform," 2024 IEEE 2nd International Conference on Electrical, Automation and Computer Engineering (ICEACE), Changchun, China, 2024, pp. 857-861, doi: 10.1109/ICEACE63551.2024.10898788.

\bibitem{ref3}
Zhou, Daming, et al. "Degradation Prediction of PEM Fuel Cell Using a Moving Window Based Hybrid Prognostic Approach." Energy, vol. 134, 2017, pp. 66-76. ScienceDirect, doi:10.1016/j.energy.2017.07.096.

\bibitem{ref4}
Chen, Kui, et al. "Aging Prognosis Model of Proton Exchange Membrane Fuel Cell in Different Operating Conditions." International Journal of Hydrogen Energy, vol. 45, no. 20, 14 Apr. 2020, pp. 11761-72. ScienceDirect, doi:10.1016/j.ijhydene.2020.02.085.

\bibitem{ref5}
Zhou, Daming, et al. "Degradation Prediction of PEM Fuel Cell Using a Moving Window Based Hybrid Prognostic Approach." Energy, vol. 134, 2017, pp. 66-76. ScienceDirect, doi:10.1016/j.energy.2017.07.096.

\bibitem{ref6}
Hua, Zhiguang, et al. "Remaining Useful Life Prediction of PEMFC Systems Based on the Multi-Input Echo State Network." Applied Energy, vol. 265, 1 May 2020, p. 114791. ScienceDirect, doi:10.1016/j.apenergy.2020.114791. 

\bibitem{ref7}
Wang Chengtao, et al."Study of dynamic half-cell potential signal of rock bolt under stray current interference based on probabilistic-based method. "Construction and Building Materials 335. 2022. doi:10.1016/J.CONBUILDMAT.2022.127481.

\bibitem{ref8}
Benchikha, Karem, et al. "Fuel Cell Ageing Prediction and Remaining Useful Life Forecasting." 2022 IEEE Vehicle Power and Propulsion Conference (VPPC), Nov. 2022. IEEE Xplore, doi:10.1109/VPPC55846.2022.10003313.

\bibitem{ref9}
Xie, Yucen, et al. "A Novel PEM Fuel Cell Remaining Useful Life Prediction Method Based on Singular Spectrum Analysis and Deep Gaussian Processes." International Journal of Hydrogen Energy, vol. 45, no. 55, 6 Nov. 2020, pp. 30942-56. ScienceDirect, doi:10.1016/j.ijhydene.2020.08.052. Accessed 24 May 2023.

\bibitem{ref10}
Le, G. T., et al. "Simulation-Informed Machine Learning Diagnostics of Solid Oxide Fuel Cell Stack with Electrochemical Impedance Spectroscopy." Journal of The Electrochemical Society, vol. 169, no. 3, 2022, p. 034530. IOP Publishing, doi:10.1149/1945-7111/ac59f4.

\bibitem{ref11}
Zhu Yuan, Smart Fuel Cell Control Unit Based on AURIX TC499, Infineon Professor Forum, Jul, 2024, Ningbo, China 

\end{thebibliography}
%

\vfill
\end{document}